\documentclass[runningheads]{llncs}
\usepackage[T1]{fontenc}
\usepackage{graphicx}
\usepackage{textcomp}
\usepackage{stfloats}
\usepackage{url}
\usepackage{verbatim}
\usepackage{graphicx}
\usepackage{cite}
\usepackage{balance}
\usepackage{algorithm}
\usepackage{algorithmic}
\usepackage{url}            
\usepackage{booktabs}       
\usepackage{amsfonts}       
\usepackage{nicefrac}       
\usepackage{microtype}      
\usepackage{bbding}
\usepackage{graphicx}
\usepackage{color}
\usepackage{amsmath,amssymb,amsfonts}
\usepackage{graphicx}
\usepackage{textcomp}
\usepackage{bm}
\usepackage{multirow}
\usepackage{booktabs}
\usepackage{bbding}
\usepackage{mfirstuc}
\usepackage[misc]{ifsym} 
\usepackage{graphics}
\usepackage{makecell}
\usepackage{colortbl} 
\definecolor{Ocean}{RGB}{129,194,234}

\usepackage[colorlinks=true,
    linkcolor=blue,
urlcolor=blue,citecolor=blue]{hyperref}

\usepackage{color}

\usepackage{xspace}
\newcommand{\mat}[1]{\boldsymbol{#1}} 
\newcommand{\ie}{\emph{i.e.}\xspace} 
\newcommand{\wrt}{\emph{w.r.t.}\xspace}

\newcommand{\methodname}{CLIP-Lung\xspace}
\begin{document}
\title{CLIP-Lung: Textual Knowledge-Guided Lung Nodule Malignancy Prediction
}
\titlerunning{CLIP-Lung for Lung Nodule Classification}
%
\author{Yiming Lei\inst{1} \and Zilong Li\inst{1 } \and Yan Shen\inst{2} \and
Junping Zhang\inst{1} \and
Hongming Shan\inst{1}}

\authorrunning{Y. Lei et al.}

\institute{Fudan University
\and 
China Pharmaceutical University
}
\maketitle              


\begin{abstract}
Lung nodule malignancy prediction has been enhanced by advanced deep-learning techniques and effective tricks. Nevertheless, current methods are mainly trained with cross-entropy loss using one-hot categorical labels, which results in difficulty in distinguishing those nodules with closer progression labels.
Interestingly, we observe that clinical text information annotated by radiologists provides us with discriminative knowledge to identify challenging samples. 
Drawing on the capability of the contrastive language-image pre-training (CLIP) model to learn generalized visual representations from text annotations, in this paper, we propose \methodname, a textual knowledge-guided framework for lung nodule malignancy prediction. 
First, \methodname introduces both class and attribute annotations into the training of the lung nodule classifier without any additional overheads in inference. 
Second, we designed a channel-wise conditional prompt (CCP) module to establish consistent relationships between learnable context prompts and specific feature maps. 
Third, we align image features with both class and attribute features via contrastive learning, rectifying false positives and false negatives in latent space.
The experimental results on the benchmark LIDC-IDRI dataset have demonstrated the superiority of \methodname, both in classification performance and interpretability of attention maps.
\end{abstract}


\section{Introduction}

Lung cancer is one of the most fatal diseases worldwide, and early diagnosis of the pulmonary nodule has been identified as an effective measure to prevent lung cancer. 
Deep learning-based methods for lung nodule classification have been widely studied in recent years~\cite{lei2020shape,liao2022learning,lei2021strided}. Usually, the malignancy prediction is often formulated as benign-malignant binary classification~\cite{lei2020shape,xie2019semi}, and the higher classification performance and explainable attention maps are impressive. Most previous works employ a learning paradigm that utilizes cross-entropy loss between predicted probability distributions and ground-truth one-hot labels. 
Furthermore, inspired by ordered labels of nodule progression, researchers have turned their attention to ordinal regression methods to evaluate the benign-unsure-malignant classification task~\cite{wu2019learning,lei2022meta,beckham2017unimodal,CORFs,liu2018ordinal}, where the training set additionally includes nodules with uncertain labels. 
Indeed, the ordinal regression-based methods are able to learn ordered manifolds and to further enhance the prediction accuracy.

\begin{figure}[t]
\centerline{\includegraphics[width=1.0\linewidth]{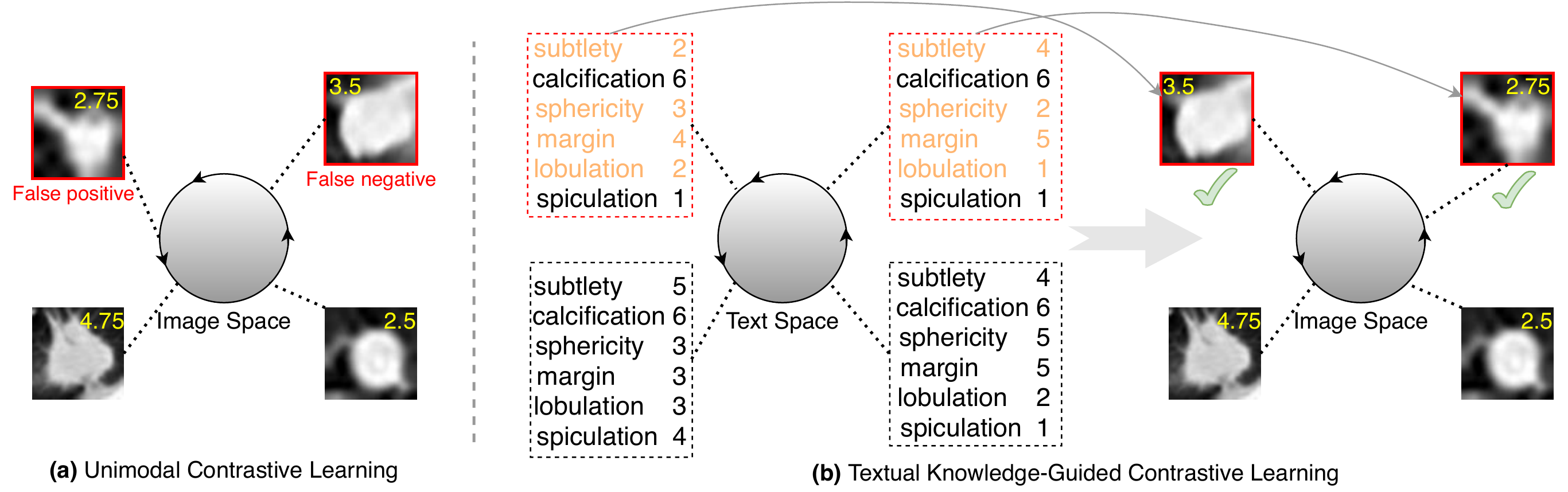}}
\caption{Motivation of \methodname. (a) Unimodal contrastive learning. (b) Proposed textual knowledge-guided contrastive learning. Yellow values are annotated malignancy scores. Dashed boxes contain pairs of textual attributes and annotated values.
}
\label{fig:motivation}
\end{figure}

However, the aforementioned methods still face challenges in distinguishing visually similar samples with adjacent rank labels.
For example in Fig.~\ref{fig:motivation} (a), since we conduct unimodal contrastive learning and map the samples onto a spherical space, the false positive nodule with a malignancy score of $2.75$ has a closer distance to that with a score of $4.75$, and the false negative one should not be closer to that of score $2.5$. 
To address this issue, we found that the text attributes, such as ``subtlety'', ``sphericity'', ``margin'', and ``lobulation'', annotated by radiologists can exhibit the differences between these hard samples. Therefore, we propose leveraging the text annotations to guide the learning of visual features. 
In reality, this also aligns with the fact that the annotated text information represents the direct justification for identifying lesion regions in the clinic.

To integrate text annotations into the image-domain learning process, an effective text encoder providing precise text features is required.
Fortunately, recent advancements in vision-language models, such as contrastive language-image pre-training (CLIP)~\cite{clip}, provide us with a powerful pre-trained text encoder learned from text-based supervisions and have shown impressive results in downstream vision tasks.
Nevertheless, it is ineffective to directly transfer CLIP to medical tasks due to the data covariate shift. Therefore, in this paper, we propose \methodname, a framework to classify lung nodules using image-text pairs. Specifically, \methodname constructs learnable text descriptions for each nodule, including class- and attribute-level. Inspired by CoCoOp~\cite{cocoop}, we proposed a channel-wise conditional prompt (CCP) module to allow nodule descriptions to guide the generation of informative feature maps. Different from CoCoOp, CCP constructs specific learnable context prompts conditioned on grouped feature maps and triggers more explainable attention maps such as Grad-CAM~\cite{grad-cam}, whereas CoCoOp provides only the common condition for all the prompt tokens. Then, we design a textual knowledge-guided contrastive learning based on obtained image features and textual features involving classes and attributes. Experimental results on LIDC-IDRI~\cite{LIDC} dataset demonstrate the effectiveness of learning with textual knowledge for improving lung nodule malignancy prediction.

The contributions of this paper are summarized as follows. 
\begin{enumerate}
    \item[1)] We proposed \methodname for lung nodule malignancy prediction which leverages clinical textual knowledge to enhance the image encoder and classifier.
    \item[2)] We designed a channel-wise conditional prompt module to establish consistent relationships among the correlated text tokens and feature maps.
    \item[3)]  We align the image features with class and attribute features through contrastive learning while generating more explainable attention maps simultaneously.
\end{enumerate}


\begin{figure}[t]
\centerline{\includegraphics[width=1.0\linewidth]{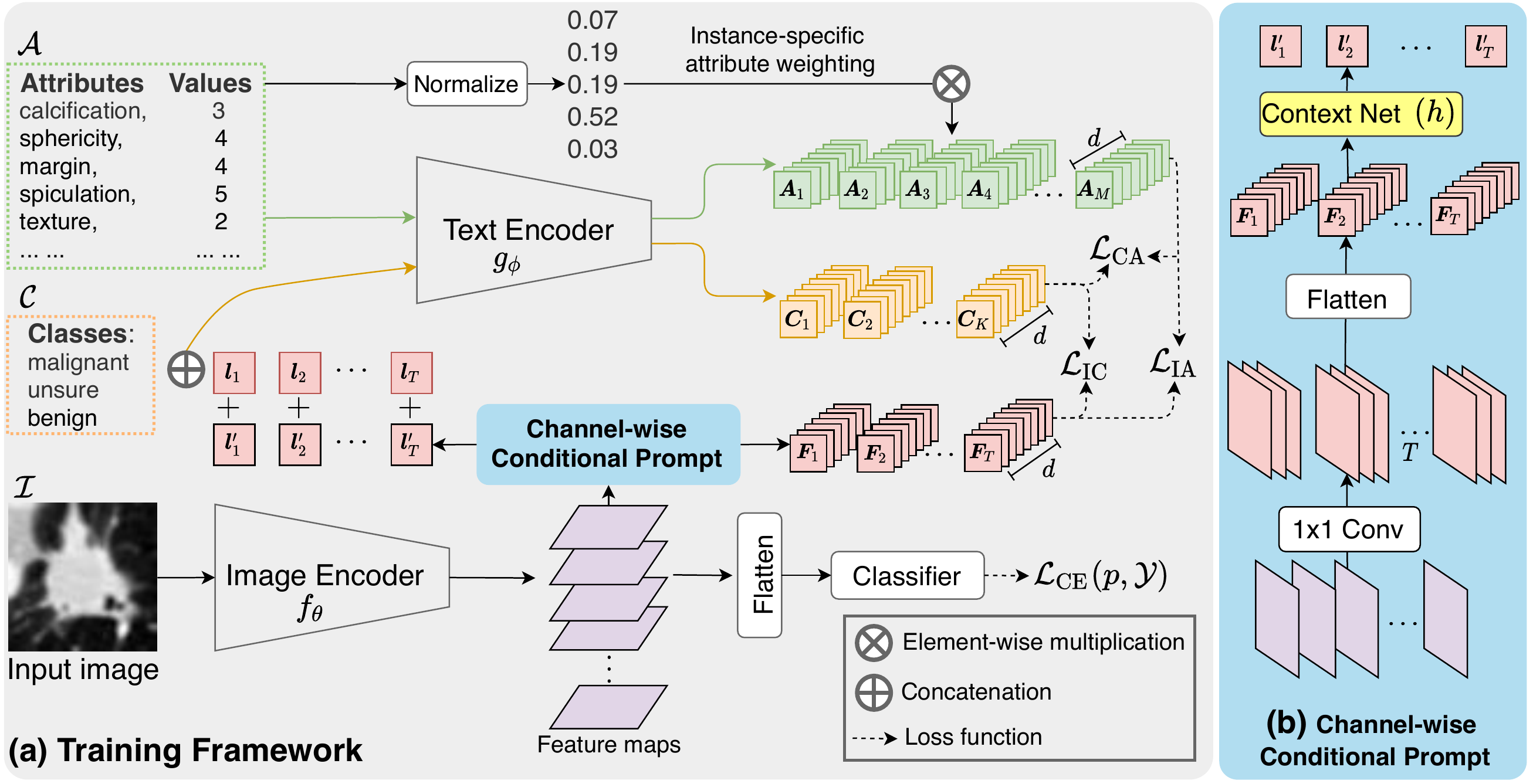}}
\caption{Illustration of the proposed \methodname.
}
\label{fig:framework}
\end{figure}

\section{Methodology}

\subsection{Overview}
\noindent\textbf{Problem formulation}\quad In this paper, we arrange the lung nodule classification dataset as $\{ \mathcal{I}, \mathcal{Y}, \mathcal{C}, \mathcal{A} \}$, where $\mathcal{I} = \{ \mat{I}_{i} \}_{i=1}^{N}$ is an image set contained $N$ lung nodule images. $\mathcal{Y} = \{ y_{i} \}_{i=1}^{N}$ is the corresponding class label set and $y_{i} \in \{1, 2, \ldots, K\}$, and $K$ is the number of classes. $\mathcal{C} = \{ \vec{c}_{k} \}_{k=1}^{K}$ is a set of text embeddings of classes. Finally, $\mathcal{A} = \{ \vec{a}_{m} \}_{m=1}^{M}$ is the set of attribute embeddings, where each element $\vec{a}_{m} \in \mathbb{R}^{d \times 1}$ is a vector representing the embedding of an attribute word such as ``spiculation''. Then, for a given sample $\{ \mat{I}_{i}, y_{i} \}$, our aim is to learn a mapping $f_{\mat{\theta}}: \mat{I}_{i} \mapsto y_{i}$, where $f$ is a deep neural network parameterized by $\mat{\theta}$. 

\noindent\textbf{\methodname}\quad In Fig.~\ref{fig:framework} (a), the training framework contains an image encoder $f_{\mat{\theta}}$ and a text encoder $g_{\mat{\phi}}$. First, the input image $\mat{I}_{i}$ is fed into $f_{\mat{\theta}}$ and then generates the feature maps. Then according to Fig.~\ref{fig:framework} (b), the feature maps are converted to channel-wise feature vectors $f_{\mat{\theta}} (\mat{I}_{i}) = \mat{F}_{t,:}$ and then to learnable tokens $\mat{l}'_{t}$. Second, we initialize the context tokens $\mat{l}_{t}$ and add them with $\mat{l}'_{t}$ to construct the learnable prompts, where $T$ is the number of context words. Next, the concatenation of the class token and $\mat{l}_{t} + \mat{l}'_{t}$ is used as input of text encoder yielding the class features $g_{\mat{\phi}} (\mat{c}_{k}) = \mat{C}_{k,:}$, note that $\mat{C}_{k,:}$ is conditioned on channel-wise feature vectors $\mat{F}_{t,:}$. Finally, the attribute tokens $\mat{a}_{m}$ are also fed into the text encoder to yield corresponding attribute features $g_{\mat{\phi}} (\mat{a}_{m}) = \mat{A}_{m,:}$. Note that the vectors $\mat{F}_{t,:}$, $\mat{l}_{t,:}$, $\mat{l}'_{t,:}$, and $\mat{C}_{k,:}$ are with the same dimension $d = 512$ in this paper. Consequently, we have image feature $\mat{F}\in \mathbb{R}^{T \times d}$, class feature $\mat{C}\in \mathbb{R}^{K \times d}$, and attribute feature $\mat{A}\in \mathbb{R}^{M \times d}$ to conduct the textual knowledge-guided contrastive learning.

\subsection{Instance-Specific Attribute Weighting}
For the attribute annotations, all the lung nodules in the LIDC-IDRI dataset are annotated with the \emph{same} eight attributes: ``subtlety'', ``internal structure'', ``calcification'', ``sphericity'', ``margin'', ``lobulation'', ``spiculation'', and ``texture''~\cite{attribute1,attribute2}, and the annotated value for each attribute is ranged from $1$ to $5$ except for ``calcification'' that is ranged from $1$ to $6$. In this paper, we fix the parameters of a pre-trained text encoder so that the generated eight text feature vectors are the same for all the nodules. Therefore, we propose an instance-specific attribute weighting scheme to distinguish different nodules. For the $i$-th sample, the weight for each $\mat{a}_{m}$ is calculated through normalizing the annotated values:
\begin{align}
w_{m} = \frac{\text{exp}(v_{m})}{\sum_{m=1}^{M}\text{exp}(v_{m})},
\label{eq:weight}
\end{align}
where $v_{m}$ denotes the annotated value of $\mat{a}_{m}$. Then the weight vector of the $i$-th sample is represented as $\vec{w}_{i} = (w_{1}, w_{2}, \ldots, w_{M})^\top \in \mathbb{R}^{M \times 1}$. Hence, the element-wise multiplication $ \vec{w}_{i} \cdot \mat{A}_{i}$ is unique to $\mat{I}_{i}$.

\subsection{Channel-wise Conditional Prompt}
CoCoOp~\cite{cocoop} firstly proposed to learn context prompts for vision-language models conditioned on visual features. However, it is inferior to align context words with partial regions of the lesion. Therefore, we propose a channel-wise conditional prompt (CCP) module, in Fig.~\ref{fig:framework} (b), to split latent feature maps into $T$ groups and then flatten them into vectors $\mat{F}_{t,:}$. Next, we denote $h(\cdot)$ as a context net that is composed of a multi-layer perceptron (MLP) with one hidden layer, and each learnable context token is now obtained by $\mat{l}'_{t} = h(\mat{F}_{t,:})$. Hence, the conditional prompt for the $t$-th token is $\mat{l}_{t} + \mat{l}'_{t}$. In addition, CCP also outputs the $\mat{F}_{t,:}$ for image-class and image-attribute contrastive learning.

\subsection{Textual Knowledge-Guided Contrastive Learning}
Contrastive learning can effectively shorten the distances between positive pairs and increase the distances between negative ones~\cite{moco,simsiam,henaff2020data}, and vision-language models also applied contrastive learning using cross-modal image-text pairs and achieved generalized image and text encoders~\cite{clip}. In \methodname, our aim is to align $\mat{F}\in \mathbb{R}^{T \times d}$ with $\mat{C} \in \mathbb{R}^{K \times d}$ and $\mat{A}\in \mathbb{R}^{M \times d}$ as illustrated in Fig.~\ref{fig:framework}, \ie, using class and attribute knowledge to regularize the feature maps.

\noindent\textbf{Image-class alignment}\quad First, the same to CLIP, we align the image and class information by minimizing the cross-entropy (CE) loss based on the prediction probability $p_{\text{i}}$:
\begin{align}
\mathcal{L}_{\text{IC}} = - \sum_{t=1}^{T} \sum_{k=1}^{K}  y_{i} \text{log} (p_{\text{i}}), \quad p_{\text{i}} = \frac{\text{exp}(\sigma(\mat{F}_{t,:}, \mat{C}_{k,:}) / \tau)}{\sum_{k'=1}^{K} \text{exp}(\sigma(\mat{F}_{t,:}, \mat{C}_{k',:}) / \tau)},  
\label{eq:IC_loss}
\end{align}
where $\mat{C}_{k,:} = g_{\mat{\phi}} (\mat{c}_{k} \bigoplus (\mat{l}_{1}+\mat{l}'_{1}, \mat{l}_{2}+\mat{l}'_{2}, \ldots, \mat{l}_{T}+\mat{l}'_{T})) \in \mathbb{R}^{d \times 1}$ and ``$\bigoplus$'' denotes concatenation, \ie, $\mat{C}_{k,:}$ is conditioned on learnable prompts $\mat{l}_{t}+\mat{l}'_{t}$. $\sigma(\cdot, \cdot)$ calculates cosine similarity and $\tau$ is temperature term. Therefore, $\mathcal{L}_{\text{IC}}$ implements the contrastive learning between channel-wise features and corresponding class features, \ie, the ensemble of grouped image-class alignment results.

\noindent\textbf{Image-attribute alignment}\quad
In addition to image-class alignment, we further expect the image features to be correlated with specific attributes. So we conduct image-attribute alignment by minimizing the InfoNCE loss~\cite{moco,clip}:
\begin{align}
\mathcal{L}_{\text{IA}} = - \sum_{t=1}^{T} \sum_{m=1}^{M} \text{log} \frac{\text{exp}(\sigma(\mat{F}_{t,:}, \mat{w}_{m,:} \cdot \mat{A}_{m,:} ) / \tau)}{\sum_{m'=1}^{M} \text{exp}(\sigma(\mat{F}_{t,:}, \mat{w}_{m',:} \cdot \mat{A}_{m',:}) / \tau)}.
\label{eq:IA_loss}
\end{align}
Due to each vector $\mat{F}_{t,:}$ is mapped from the $t$-th group of feature maps through context net $h(\cdot)$, then $\mathcal{L}_{\text{IA}}$ indicates which attribute the $\mat{F}_{t,:}$ is closest to. Therefore, certain feature maps can be guided by specific annotated attributes.

\noindent\textbf{Class-attribute alignment}\quad
Although the image features have been aligned with classes and attributes, the class embeddings obtained by the pre-trained CLIP encoder may shift in latent space. This will result in inconsistent class space and attribute space, \ie, annotated attributes do not match the corresponding classes, which is contradictory to the actual clinical diagnosis. To avoid this weakness, we further align the class and attribute features:
\begin{align}
\mathcal{L}_{\text{CA}} = - \sum_{k=1}^{K} \sum_{m=1}^{M} \text{log} \frac{\text{exp}(\sigma(\mat{C}_{k,:}, \mat{w}_{m,:} \cdot \mat{A}_{m,:} ) / \tau)}{\sum_{m'=1}^{M} \text{exp}(\sigma(\mat{C}_{k,:}, \mat{w}_{m',:} \cdot \mat{A}_{m',:}) / \tau)},
\label{eq:CA_loss}
\end{align}
and this loss implies semantic consistency between classes and attributes. 

Finally, the total loss function is defined as follows:
\begin{align}
\mathcal{L} = \mathbb{E}_{\mat{I}_{i} \in \mathcal{I}} \big[ \mathcal{L}_{\text{CE}} + \mathcal{L}_{\text{IC}} + \alpha \cdot \mathcal{L}_{\text{IA}} + \beta \cdot \mathcal{L}_{\text{CA}} \big],
\label{eq:total_loss}
\end{align}
where $\alpha$ and $\beta$ are hyperparameters for adjusting the losses and are set as $1$ and $0.5$, respectively. $\mathcal{L}_{\text{CE}}$ denotes the cross-entropy loss between predicted probabilities obtained by the classifier and the ground-truth labels.
Note that during the inference phase, test images are only fed into the trained image encoder and classifier, therefore, \methodname does not introduce any additional  computational overhead in inference.


\section{Experiments}

\subsection{Dataset and Implementation Details}
\noindent\textbf{Dataset}\quad LIDC-IDRI~\footnote{\url{https://wiki.cancerimagingarchive.net/pages/viewpage.action?pageId=1966254}} is a dataset for pulmonary nodule classification or detection based on low-dose CT, which involves 1,010 patients. All the nodules were labeled with scores from $1$ to $5$, indicating the malignancy progression. We cropped all the nodules with a square shape of a doubled equivalent diameter at the annotated center, then resized samples to the volume of $32 \times 32 \times 32$. Following~\cite{lei2020shape,lei2022meta}, we modified the first layer of the image encoder to be with $32$ channels. According to existing works~\cite{lei2022meta,wu2019learning}, we regard a nodule with an average score between $2.5$ and $3.5$ as unsure nodules, benign and malignant categories are those with scores lower than $2.5$ and larger than $3.5$, respectively. In this paper, we construct three sub-datasets: LIDC-A contains three classes of nodules both in training and test sets; according to~\cite{lei2022meta}, we construct the LIDC-B, which contains three classes of nodules \emph{only} in the training set, and the test set contains benign and malignant nodules; LIDC-C includes benign and malignant nodules both in training and test sets.

\noindent\textbf{Experimental settings}\quad In this paper, we apply the CLIP pre-trained text encoder ViT-B/16 as the text encoder for \methodname, and the image encoder we used is ResNet-18~\cite{ResNet} due to the relatively smaller scale of training data. The image encoder is initialized randomly. Note that for the text branch, we froze the parameters of the text encoder, and update the learnable tokens $\vec{l}$ and $\vec{l}'$ during training. The learning rate is $0.001$ following the cosine decay, the optimizer is stochastic gradient descent with momentum $0.9$ and weight decay $0.00005$. The temperature $\tau$ is initialized as $0.07$ and updated during training.
All of our experiments are implemented with PyTorch~\cite{pytorch} and trained with NVIDIA A100 GPUs. The experimental results are reported with average values through five randomly independent split folds. For different classes, we report the recall and F1-score values, and ``$\pm$'' indicates standard deviation.

\subsection{Experimental Results and Analysis}
\noindent\textbf{Performance comparisons}\quad In Table~\ref{tab:train3_test3}, we compare the classification performances on the LIDC-A dataset, where we regard the benign-unsure-malignant as an ordinal relationship. Compared with ordinal classification methods such as Poisson, NSB, UDM, and CORF, \methodname achieves the highest accuracy and F1-scores for the three classes, which demonstrates the effectiveness of textual knowledge-guided learning. CLIP and CoCoOp also outperform ordinal classification methods and show the superiority of large-scale pre-trained text encoders. Furthermore, \methodname obtained higher recalls than CLIP and CoCoOp \wrt benign and malignant classes, however, the recall of unsure is lower than theirs, we argue that this is due to the indistinguishable textual annotations such as similar attributes of different nodules.

In Table~\ref{tab:train32_test2}, we compare the performances on LIDC-B and LIDC-C datasets. \methodname obtains higher evaluation values other than recalls of benign class. We conjecture the reason is that most of the benign nodules are with similar appearances and subtle differences in text attributes, therefore, aligning these two types of features is difficult and the text features will be biased to those of malignant nodules. 

\begin{table*}[t]
\caption{Classification results on the test set of LIDC-A.}
\centering
\begin{tabular*}{1.0\linewidth}{@{\extracolsep{\fill}}l*{7}{c}}
\toprule
\multirow{2}{*}{Method} & \multirow{2}{*}{Accuracy} & \multicolumn{2}{c}{Benign} & \multicolumn{2}{c}{Malignant} & \multicolumn{2}{c}{Unsure} \\ \cline{3-4} \cline{5-6} \cline{7-8}
  &  & Recall & F1 & Recall & F1 & Recall & F1 \\
\midrule

CE Loss & 54.2$\pm0.6$ & 72.2 & 62.0 & 64.4 & 61.3 & 29.0 & 36.6 \\
Poisson~\cite{beckham2017unimodal} & 52.7$\pm0.7$ & 60.5 & 56.8 & 58.4 & 58.7 & 41.0 & 44.1 \\
NSB~\cite{liu2018ordinal} & 53.4$\pm0.7$ & \bf{80.7} & 63.0 & \bf{67.3} & 63.8 & 16.0 & 24.2 \\
UDM~\cite{wu2019learning} & 54.6$\pm0.4$ & 76.7 & 64.3 & 49.5 & 53.5 & 32.5 & 39.5 \\
CORF~\cite{CORFs} & 56.8$\pm0.4$ & 71.3 & 63.3 & 61.3 & 62.3 & 38.5 & 44.3 \\
CLIP~\cite{clip} & 56.6$\pm0.3$ & 59.5 & 59.2 & 55.2 & 60.0 & 53.9 & 52.2 \\
CoCoOp~\cite{cocoop} & 56.8$\pm0.6$ & 59.0 & 59.2 & 55.2 & 60.0 & \bf{55.1} & 52.8 \\
\bf{\methodname} & \bf{60.9}$\pm0.4$ & 67.5 & \bf{64.4} & 60.9 & \bf{66.3} & 53.4 & \bf{54.1} \\

\bottomrule
\end{tabular*}
\label{tab:train3_test3}
\end{table*}

\begin{table*}[t]
\caption{Classification results on test sets of LIDC-B and LIDC-C.}
\centering
\begin{tabular*}{1.0\textwidth}{@{\extracolsep{\fill}}l*{5}{c}*{5}{c}}
\toprule
& \multicolumn{5}{c}{LIDC-B} & \multicolumn{4}{c}{LIDC-C} \\ \cline{2-6} \cline{7-11}
\multirow{2}{*}{Method} & \multirow{2}{*}{Accuracy} & \multicolumn{2}{c}{Benign} & \multicolumn{2}{c}{Malignant} & \multirow{2}{*}{Accuracy} & \multicolumn{2}{c}{Benign} & \multicolumn{2}{c}{Malignant}\\ \cline{3-6}\cline{8-11}
 &  & Recall & F1  & Recall & F1  &  & Recall & F1  & Recall & F1  \\
\midrule

CE Loss & 83.3$\pm0.6$ & 92.4 & 88.4 & 63.4 & 70.3 & 85.5$\pm0.5$ & 91.5 & 89.7 & 72.3 & 75.6\\
Poisson~\cite{beckham2017unimodal} & 81.8$\pm0.4$ & 94.2 & 87.7 & 54.5 & 65.1 & 84.0$\pm0.3$ & 87.9 & 88.3 & 75.2 & 74.5\\
NSB~\cite{liu2018ordinal} & 78.1$\pm0.5$ & 90.6 & 85.8 & 50.5 & 60.7 & 84.9$\pm0.7$ & 91.0 & 89.2 & 71.3 & 74.6\\
UDM~\cite{wu2019learning} & 79.3$\pm0.4$ & 87.0 & 86.2 & 62.4 & 67.7 & 84.6$\pm0.5$ & 88.8 & 88.8 & 75.2 & 75.2 \\
CORF~\cite{CORFs} & 81.5$\pm0.3$ & \bf{95.9} & 87.8 & 49.5 & 62.8 & 83.0$\pm0.2$ & 87.9 & 87.7 & 72.3 & 72.6\\
CLIP~\cite{clip} & 83.6$\pm0.6$ & 92.0 & 88.7 & 64.4 & 70.4 & 87.5$\pm0.3$ & 92.0 & 91.0 & 77.0 & 78.8 \\
CoCoOp~\cite{cocoop} & 86.8$\pm0.7$ & 94.5 & 90.9& 69.0 & 75.9 & 88.2$\pm0.6$ & \bf{95.0} & 91.8 & 72.4 & 78.8 \\
\bf{\methodname} & \bf{87.5}$\pm0.3$ &  94.5 & \bf{91.7} & \bf{72.3} & \bf{79.0}  & \bf{89.5}$\pm0.4$ & 94.0 & \bf{92.8} & \bf{80.5} & \bf{82.8} \\

\bottomrule
\end{tabular*}
\label{tab:train32_test2}
\end{table*}

\begin{figure}
\centerline{\includegraphics[width=1.0\linewidth]{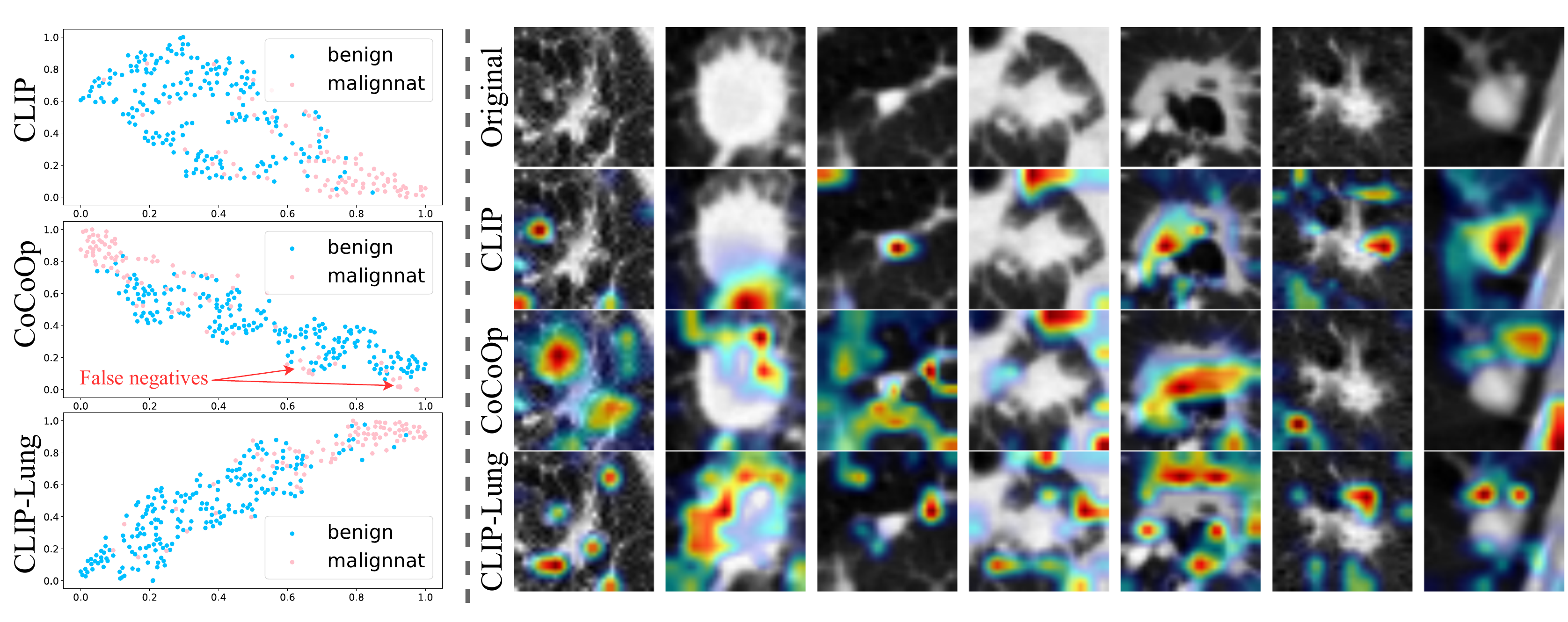}}
\caption{The t-SNE (\textbf{Left}) and Grad-CAM (\textbf{Right}) results.
}
\label{fig:visualization}
\end{figure}

\noindent\textbf{Visual features and attention maps}\quad
To illustrate the influence of incorporating class and attribute knowledge, we provide the t-SNE~\cite{tsne} Grad-CAM~\cite{grad-cam} results obtained by CLIP, CoCoOp, and \methodname. In Fig.~\ref{fig:visualization}, we can see that CLIP yields a non-compact latent space for two kinds of nodules. CoCoOp and \methodname alleviate this phenomenon, which demonstrates that the learnable prompts guided by nodule classes are more effective than fixed prompt engineering. Further, compared with \methodname, CoCoOp could not consider the attribute information to learn the prompts, therefore, it results in more false negatives in latent space. From the attention maps we can observe that CLIP cannot precisely capture spiculation and lobulation regions that are highly correlated with malignancy. Simultaneously,  \methodname performs better than CoCoOp, which demonstrates the guidance from textual descriptions such as ``spiculation''.

\noindent\textbf{Ablation studies}\quad
In Fig.~\ref{tab:ablation}, we verify the effectiveness of different loss components on the three constructed datasets. Based on $\mathcal{L}_{\text{IC}}$, $\mathcal{L}_{\text{IA}}$ and $\mathcal{L}_{\text{CA}}$ improve the performances on LIDC-A, indicating the effectiveness of capturing fine-grained features of ordinal ranks using class and attribute texts. However, they perform relatively worse on LIDC-B and LIDC-C, especially the $\mathcal{L}_{\text{IC}} + \mathcal{L}_{\text{CA}}$. That is to say, $\mathcal{L}_{\text{IA}}$ is more important in latent space rectification, \ie, image-attribute consistency. In addition, we observe that $\mathcal{L}_{\text{IC}} + \mathcal{L}_{\text{IA}}$ performs better than $\mathcal{L}_{\text{IA}} + \mathcal{L}_{\text{CA}}$, which is attributed to that $\mathcal{L}_{\text{CA}}$ regularizes the image features indirectly.
\begin{table}[!t]
\caption{Ablation study on different losses. We report classification accuracies.
}
\label{tab:ablation}
\centering
\begin{tabular*}{1.0\linewidth}{@{\extracolsep{\fill}}cccccc}

\toprule
$\mathcal{L}_{\text{IC}}$ & $\mathcal{L}_{\text{IA}}$ & $\mathcal{L}_{\text{CA}}$ & LIDC-A & LIDC-B & LIDC-C \\

\midrule
\checkmark & \ &                     & 56.8$\pm0.6$ & 86.8$\pm0.7$ & 88.2$\pm0.6$ \\
\checkmark & \checkmark &            & 59.4$\pm0.4$ & 86.8$\pm0.6$ & 86.7$\pm0.4$ \\
           & \checkmark & \checkmark & 58.1$\pm0.2$ & 85.7$\pm0.6$ & 87.5$\pm0.5$ \\
\checkmark &            & \checkmark & 56.9$\pm0.3$ & 84.7$\pm0.4$ & 84.0$\pm0.7$ \\
\checkmark & \checkmark & \checkmark & 60.9$\pm0.4$ & 87.5$\pm0.5$ & 89.5$\pm0.4$ \\

\bottomrule
\end{tabular*}
\vspace{-5mm}
\end{table}


\section{Conclusion}

In this paper, we proposed a textual knowledge-guided framework for pulmonary classification, named \methodname. We explored the utilization of clinical textual annotations based on large-scale pre-trained text encoders. \methodname aligned the different modalities of features generated from nodule classes, attributes, and images through contrastive learning. 
Most importantly, \methodname establishes correlations between learnable prompt tokens and feature maps using the proposed CCP module, and this guarantees explainable attention maps localizing fine-grained clinical features. 
Finally, \methodname outperforms compared methods, including CLIP on LIDC-IDRI benchmark. 
Future work can concentrate on extending \methodname with more diverse textual knowledge.
%
%
%

%
%
%
%

\end{document}